\documentclass[runningheads]{llncs}
\usepackage{amsmath} 
\usepackage{algorithm}          
\usepackage{algorithmic}        
\usepackage{graphicx}           
\usepackage{subcaption}         %
\usepackage{multirow}           
\usepackage{csquotes}           
\usepackage{makecell}           
\usepackage[table]{xcolor}      
\usepackage{commenting}         
\usepackage{comment}            
\usepackage{paralist}           

\title{Details Make a Difference: Object State-Sensitive Neurorobotic Task Planning\thanks{Published in the Proceedings of the International Conference on Artificial Neural Networks, 2024.}} 
\titlerunning{Object State-Sensitive Agent} %
\author{Xiaowen Sun\thanks{Corresponding author}\and
Xufeng Zhao \and
Jae Hee Lee\and
Wenhao Lu\and \\
Matthias Kerzel\and
Stefan Wermter}
\authorrunning{X. Sun et al.}
\institute{Knowledge Technology, Department of Informatics, University of Hamburg\\
\email{\{xiaowen.sun, xufeng.zhao, jae.hee.lee, wenhao.lu,\\ matthias.kerzel, stefan.wermter\}@uni-hamburg.de}
\newline \url{www.knowledge-technology.info}}

\begin{document}

\maketitle
\begin{abstract}
The state of an object reflects its current status or condition and is important for a robot's task planning and manipulation. However, detecting an object's state and generating a state-sensitive plan for robots is challenging. Recently, pre-trained Large Language Models (LLMs) and Vision-Language Models (VLMs) have shown impressive capabilities in generating plans. However, to the best of our knowledge, there is hardly any investigation on whether LLMs or VLMs can also generate object state-sensitive plans. To study this, we introduce an Object State-Sensitive Agent (OSSA), a task-planning agent empowered by pre-trained neural networks. We propose two methods for OSSA: (i) a modular model consisting of a pre-trained vision processing module (dense captioning model, DCM) and a natural language processing model (LLM), and (ii) a monolithic model consisting only of a VLM. To quantitatively evaluate the performances of the two methods, we use tabletop scenarios where the task is to clear the table.  We contribute a multimodal benchmark dataset that takes object states into consideration. Our results show that both methods can be used for object state-sensitive tasks, but the monolithic approach outperforms the modular approach. 
The code for OSSA is available at \url{https://github.com/Xiao-wen-Sun/OSSA}
\keywords{Object State Identification \and Artificial Intelligence  \and Robotics \and Language Models \and Multimodality}

\end{abstract}

\section{Introduction}
Object attributes such as color, shape, and size play important roles in shaping the diverse states objects can exhibit, significantly impacting our daily interactions. 
Understanding and managing these states is crucial as overlooking them can result in unforeseen consequences. 
Humans intuitively rely on their understanding of object states and common sense to interact with everyday objects. 
However, integrating state-sensitive knowledge poses a significant challenge for robotic systems. 
Enumerating and seamlessly integrating this nuanced understanding into robotic systems remains a complex endeavor.

Recent approaches that leverage Large Language Models (LLMs)~\cite{zhao2023survey} have shown promising results in tasks that require human-level commonsense knowledge. Such approaches use LLMs' commonsense reasoning ability to interpret natural language goals~\cite{zhao2024large}, 
such as `put one apple into the fridge.' According to those goals, LLM generates the suggested action plan, `next actions: pick up the apple, move to the kitchen, open the fridge, \dots.'
However, the object's physical state (e.g., intact apple, sliced apple, etc.) is crucial yet less considered in the task planning
~\cite{singh2023progprompt}.
To the best of our knowledge, there is hardly any research that leverages a pre-trained neural network (e.g., LLM or VLM) to address the integration of object states into planning tasks for household robots.
To address this gap, we introduce an \textbf{O}bject \textbf{S}tate-\textbf{S}ensitive \textbf{A}gent (\textbf{OSSA}), which utilizes commonsense reasoning of pre-trained neural networks for robot task planning.

We pose several real-world challenges in object state-sensitive agent (OSSA) task planning. 
First, in order to solve the tasks, the agent does not only involve identifying different objects in the scene but also distinguishes between their \emph{states}. 
For example, in the `clear the table'~\cite{nyga_grounding_2018,gutman_evaluating_2023} task, a robot needs to be able to distinguish between \emph{whole} and \emph{sliced} fruit, and between \emph{clean} and \emph{dirty} plates. 
This is a challenge because existing state-of-the-art object detection models often fail at differentiating between objects in different states. A plan lacking state-sensitive awareness may lead to unexpected results.

Second, the agent needs to employ commonsense reasoning for taking \emph{state-sensitive} actions that correspond to the object states of various scenarios instead of asking the users for an exhaustive design or an intervention. 
For example, whole fruit go directly into the fridge or to the cupboard, while sliced fruit can either go into the fridge or be discarded into the trash bin if they are regarded as leftovers.
One way to solve the commonsense reasoning problem is using rule-based symbolic approaches~\cite{nyga_grounding_2018}, but by design, they do not generalize to situations outside their rule base, i.e., they cannot handle new objects and states. Commonsense reasoning with a data-driven model trained on a large dataset that generalizes well (e.g., a large language model~\cite{wang2024large}) is, therefore, a more viable choice.

Third, the robot needs to identify cases where common sense should not dominate. 
For example, a robot has to take into account the preferences of the user when handling specific objects in specific states~\cite{wu2023tidybot}. 
In the table clearing example, different users might handle the leftover food differently, e.g., some might prefer to discard the leftovers, while others are more frugal and prefer to keep the leftovers for the next meal. The robot, therefore, needs to detect situations where different user preferences come into play, and has to ask the user for clarification instead of arbitrarily choosing one on its own. Existing approaches, however, do not take user preferences into account~\cite{sun2022learning,ren2023robots}.

\begin{figure}[t]
    \centering
    \includegraphics[page=1,width=1\linewidth]{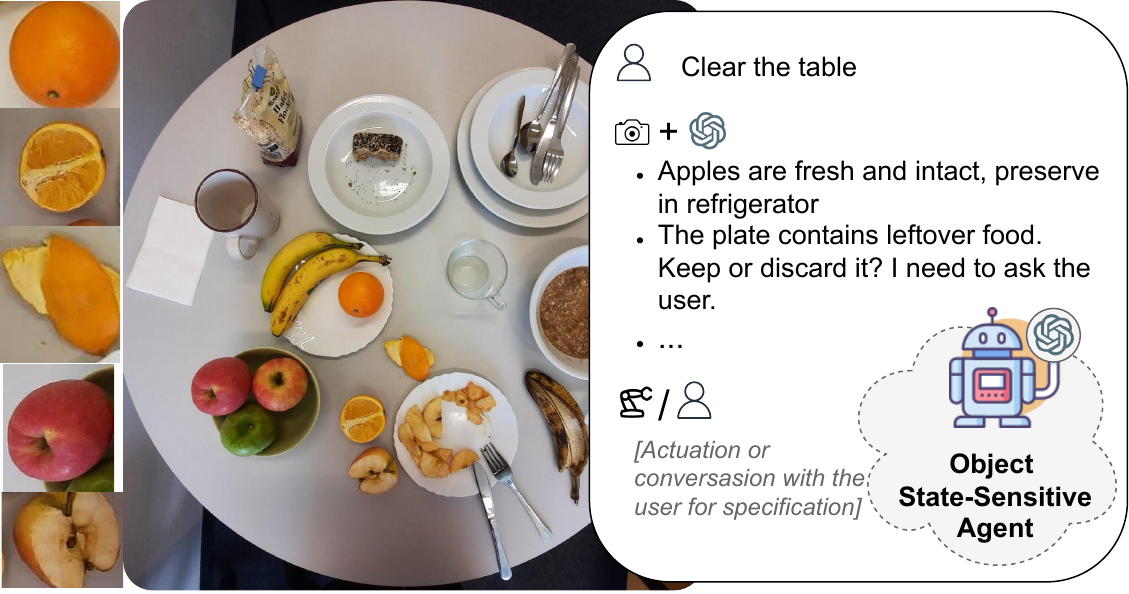} 
    \caption{
      The given scene contains various objects in various states. For example, orange, half-orange, and orange peel; clean napkin and dirty napkin; banana and banana peel. Based on commonsense knowledge, the agent sorts the objects (discard the banana peel in the trash bin; keep the bananas in the cupboard). However, the robot is not able to decide how to deal with the leftover food because different people may have different preferences regarding leftover food (e.g., half orange and half bread).}
    \label{fig: example scene}
\end{figure}

In this paper, we study the problem of state-sensitive instruction following. We investigate two different methods, (i) a modular model consisting of an object detection module and an LLM, and (ii) a monolithic VLM-only model to caption and reason universally. We use cluttered tabletop scenarios (in Fig.~\ref{fig: example scene}) as a specific application of our approaches, where the task is to clear the table based on the object's attributes, states, and user preferences. We summarize our contributions as follows:
\begin{compactenum}
    \item We introduce an object state-sensitive agent (OSSA) that can perceive fine-grained scene information (objects and their states) and generate an appropriate object manipulation plan for the robot's low-level executor to fulfill a given task.
    \item To explore a pre-trained neural network's capability for object state-sensitive task planning, we propose two methods for OSSA: a modular approach consisting of dense captioning model (DCM) and LLM modules and a monolithic approach consisting only of a VLM.
    \item To evaluate the proposed methods, we formulate an instruction-following task for robots that focuses on the objects' attributes and states in a tabletop scenario.
    \item We provide an open-source benchmark dataset containing various object states, which involves 40 scenarios and 184 objects in total.
    
\end{compactenum}

\section{Related Work}

\subsection{Vision and Language Models for Task Planning in Robotics}
Recently, vision-language models (VLMs)~\cite{wang2024large,wang2023large} and large language models (LLMs)~\cite{zhao2023survey} have transformed robot task planning, with the integration of LLMs' commonsense knowledge into planning for embodied agents garnering growing interest~\cite{WPE05,huang2022inner,brohan2023can,song2023llm,rana2023sayplan,Zhao23ChatEnvironment,singh2023progprompt,zhao2024large}.

However, these methods assume a singular state for objects in their plans, which may not suit the variability of the domestic settings we study. 
The most high-performance perception frameworks~(e.g., OWL-ViT~\cite{minderer2205simple}, OWL-V2~\cite{minderer2024scaling}, YOLO~\cite{chen2022you}, CLIP~\cite{radford2021learning}, etc.) that the existing work is using~\cite{zhou2023language} are not trained to distinguish the object states, e.g., half apples, apple, dirty plate, and clean plate.
Thence, we conclude that these approaches are not sensitive to object states.

One of the closely related works to ours is TidyBot~\cite{wu2023tidybot}, a system that enables a domestic cleaning robot to adapt to individual user preferences in managing objects, recognizing that each item may require a unique approach based on personal taste. They leverage the summarization capabilities of LLMs to get the user's preference for sorting the objects, e.g., `put light-colored clothes in the drawer and dark-colored clothes in the closet'. In our work, we take advantage of commonsense knowledge from LLMs to sort the objects and detect when human preference needs to be involved in automation tasks.

In other research, VILA~\cite{hu2023look} utilizes GPT-4V(vision) as VLM to do task planning, which is also close to our work. However, the difference is that they focus on long-horizon robotic tasks, where the manipulation actions and goals can be derived from the user instruction. 
In our work, we focus on automation tasks where the robot manipulation goals stem from visual perception. 
Then our model performs reasoning and generates a manipulation plan, according to the visual perception.

\subsection{Benchmarks for Task Planning in Household Robots}
Gutman et al.~\cite{gutman_evaluating_2023} use the \texttt{clear the table} as an experiment task to explore the users' preference for different autonomy levels of an assistive robot. The results show that the participants prefer the highly autonomous assistant. Other research uses \texttt{table clearing task} to study the robot to understand and execute humans' natural language instructions~\cite{nyga_grounding_2018}. 
However, overall, their work does not mention that the objects from the same category may be in a different state, which would change the robot's action.

From the computer vision research perspective, two datasets~\cite{jelodar2018identifying} focus on object state recognition in cooking, e.g., whole, peeled, floured, etc. The Object State Detection Dataset~\cite{gouidis2022} involves the object states: open, close, empty, containing something liquid, containing something solid, plugged, unplugged, folded, and unfolded. However, it does not consider the leftover food in our daily life.
From the robot kinematics research perspective, many benchmarks~\cite{brohan2022rt,zitkovich2023rt,jang2022bc,valmeekam2024planbench} and methods have been proposed to enable robots to complete certain manipulation tasks, e.g., picking, wiping, dragging, pushing, pouring, and placing. 
We focus on automated task planning and research how to utilize these enabled robot skills to complete certain complex tasks.
For example, the robot, according to various object states, generates plans for a robot low-level executor that receives a specification of how to `pick' and where to `place' the target.

\section{Methodology}
\subsection{Architecture}\label{sec: system}
We introduce an Object State-Sensitive Agent (OSSA) that can perceive fine-grained scene information (object name and state, etc.) and generate an appropriate object manipulation plan for the robot's low-level executor according to those visual perceptions. The pseudocode of the system architecture is provided in Algorithm~\ref{alg: system architecture}.
OSSA generates object manipulation plans when an utterance ($U$) and an image of the table ($I$) are given.
A chat system is for ambiguous situations. If uncertainties arise (e.g., encountering a peeled pear on the table), the robot will ask the user for disambiguation. Lastly, a low-level executor will execute language-conditioned instructions, such as picking and placing for fulfilling the tasks. 
We assume that the low-level executor has these skills, which can be acquired via machine learning~\cite{ozdemir2022language} approaches (e.g., reinforcement learning~\cite{kalashnikov2018scalable,beik2019mixed} and imitation learning~\cite{jang2022bc}) or direct hard-coding.

\begin{algorithm}[t]
    \caption{System architecture}
    \label{alg: system architecture}
    \begin{algorithmic}[1] 
        \STATE $r \leftarrow \text{Robot.Initialize}()$ 
        \STATE $commands = NULL$ 
        \WHILE{True}{ 
            \STATE $U \leftarrow \text{r.GetUserUtterance()}$ 
            \STATE $I \leftarrow \text{r.GetImageofTable()}$ 
            \STATE $ \{ \text{OMP}_1, \dots, \text{OMP}_n \} \leftarrow \text{r.\textbf{OSSA}}(U, I)$
            \FOR{$\text{OMP}_i$ in $\{ \text{OMP}_1, \dots, \text{OMP}_n \} $}
                \IF{$\text{OMP}_i.destination$ is "uncertain"}
                    \STATE $AI_{Response} \leftarrow \text{r.ChatSystem}(\text{OMP}_i)$
                    \STATE $U \leftarrow \text{r.GetUserUtterance()}$ 
                    \STATE $\text{OMP}_i \leftarrow \text{r.ChatSystem}(U)$ \hfill \# \text{ChatSystem revise} $\text{OMP}_{i}$
                    \STATE $commands.append(\text{OMP}_i)$
                \ELSE 
                    \STATE $commands.append(\text{OMP}_i)$
                \ENDIF 
            \ENDFOR 
        \IF{$commands$ not NULL} 
            \STATE $\text{r.LowLevelExecutor(commands)}$ 
            \STATE $commands = NULL$ 
        \ENDIF } 
        \ENDWHILE
    \end{algorithmic}
\end{algorithm}

\subsection{Object Manipulation Plan Generation} \label{sec: object manipulation plan}
\begin{figure}[t]
    \centering
    \begin{subfigure}[b]{0.9\linewidth}
        \centering
        \fbox{\includegraphics[width=0.9\linewidth]{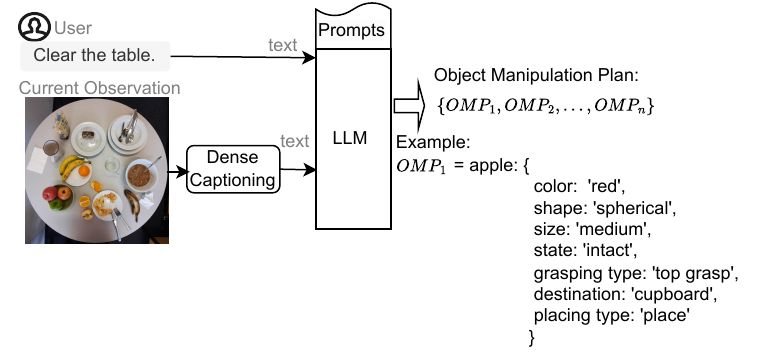}}
        \caption{Modular Method: OSSA-LLM-DCM.}
        \label{fig: method llm}
    \end{subfigure}\\
    \begin{subfigure}[b]{0.9\linewidth}
        \centering
        \fbox{\includegraphics[width=0.9\linewidth]{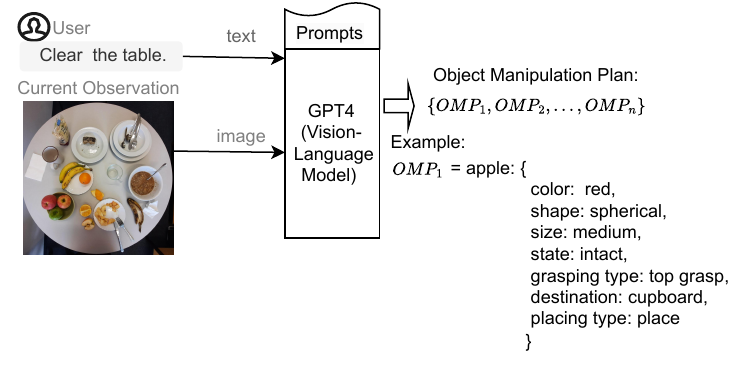}}
        \caption{Monolithic Method: OSSA-VLM}
        \label{fig: method vlm}
    \end{subfigure}
    \caption{Overview of our two proposed methods for OSSA: (a) OSSA-LLM-DCM represents the modular model that combines a prompt large language model (LLM) and a dense captioning model (DCM); (b) OSSA-VLM represents only a vision-language model (VLM).}
    \label{fig: methods}
\end{figure}

To explore the best way to employ commonsense knowledge~\cite{zhao2024large} and reasoning~\cite{huang2022inner} capabilities of pre-trained neural networks (e.g., LLMs or VLMs) for object sensitive-state agent task planning (OSSA), we propose two categories of methods for object manipulation plan generation. The input of the method is a scene ($I$) and a user’s utterance ($U$).
We prompt LLMs to generate structured responses in a JSON-like format, `object name':~\{ `color', `size', `shape', `container', `state', `grasping type', `destination', `placing type'~\}. 
`Color', `size', and `shape' are the object attributes that are visible in the scene. 
The object's physical `state' is visible in the scene. However, reasoning the human-defined object state requires commonsense knowledge.
`Grasping type' is an action that informs the robot how to grasp the target at the initial place. 
`Destination' is a place that informs the robot where the target should be stored. 
`Placing type' is an action that informs the robot what it should do at the destination. 
Those three items are from commonsense knowledge reasoning.
For object state detection, a computer vision system dense captioning~\cite{johnson2016densecap} can localize salient regions in images and use natural language to describe them. The natural language description of a salient region includes the condition of the object in the region. 
Therefore, state-of-the-art dense captioning can be utilized in our study. 
However, the output of the dense captioning model does not include `grasping type', `placing type', and `destination'; another module with commonsense knowledge and reasoning abilities is needed to generate them.
GPT-4~\cite{yang2023dawn,openai} can not only be a visual perception model but also perform as a vision-language model.
In this study, we utilize both of the abilities of GPT-4. 
We propose two methods: i) a modular model consisting of a vision processing module (dense captioning model) and natural language processing model (LLM), and (ii) a monolithic VLM-based model.

\subsubsection{Modular Method}\label{sec: lLM-based} A modular pipeline~\cite{brohan2023can,Zhao23ChatEnvironment,rana2023sayplan,song2023llm,singh2023progprompt,huang2022inner,wu2023tidybot}
combines two models, e.g., vision and language. First, a vision detection model (e.g., YOLO~\cite{chen2022you}, ViLD~\cite{gu2021open}, OWL-ViT~\cite{minderer2205simple}, OWL-V2~\cite{minderer2024scaling}, etc.) detects the objects in the scene. Second, a large language model processes the user's instruction and the output of the vision model. 
Those vision models are trained to detect the abstract object category.
Therefore, we cannot use their methods directly.

The dense captioning model does not only localize salient regions in images but also uses natural language to describe them~\cite{johnson2016densecap}.
Instead of object detection models,
we use a dense captioning model to get the description of salient regions of a scene as text. As Fig.~\ref{fig: method llm} shows, a prompt LLM processes a user instruction and dense caption. 
The quality of the dense captions is an important factor influencing the model's performance. 
We choose the state-of-the-art dense captioning model GRiT~\cite{wu2022grit}. 
Additionally, GPT-4V was also successfully used for image or video understanding~\cite{lin2023mm,wang2024large,wake2023gpt}. 
In this work, we use both GPT-4V and GRiT to generate the dense caption for a given image.

\begin{figure}[t]
    \centering
    \fbox{\includegraphics[width=0.95\linewidth]{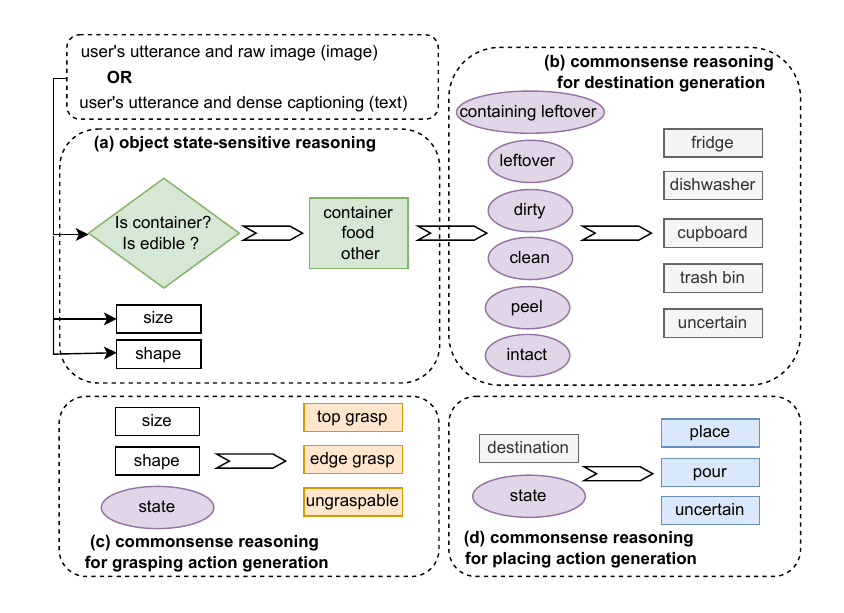}}
    \caption{Chain-of-thought for OSSA. 
    (a) The pre-trained model (e.g., LLM or VLM) utilizes commonsense knowledge to reason about the object state;
    (b) according to the object's state and user's preference, the model generates a destination for the object;
    (c) according to the object's state, shape, and size, the model generates a grasping action for the object;
    (d) according to the object's state and destination, the model generates a placing action for the object.}
    \label{fig: cot for ossa}
\end{figure}

\subsubsection{Monolithic Method}\label{sec: VLM-based} As shown in Fig.~\ref{fig: method vlm}, the user's instruction $U$ and image of a table $I$ are the input of a monolithic model. The model's output is object manipulation plans: $\{\text{OMP}_1, \text{OMP}_2, \dots, \text{OMP}_n \}$.
Yang et al.~\cite{yang2023dawn,openai} are given preliminary explorations with GPT-4V(ision). GPT-4V has also shown remarkable results for game action prediction in the simulation environment~\cite{gong2023mindagent,durante2024agent}. The approach VILA by Hu et al. ~\cite{hu2023look} unveils the capability of GPT-4V for long-horizon robotic planning to generate a sequence of actionable steps. The difference from those works is that we leverage GPT-4V to recognize the object state and generate a manipulation plan for an object according to its state.

\subsubsection{Prompts for Object State Detection and Reasoning}\label{sec: object states detection and Reasoning}
LLMs and VLMs are pre-trained on massive datasets from a diverse set of sources, which allows them to acquire human-level commonsense knowledge for handling object states. To unleash their full potential, however, it is necessary to devise appropriate prompts and guide them; otherwise, their generation cannot be directly used for robot tasks~\cite{brohan2023can,ren2023robots}. In this study, we utilize an LLM or a VLM as a high-level planner for the robot's low-level executor. Fig.~\ref{fig: cot for ossa} shows a schematic chart of the chain-of-thought~\cite{wei2022chain,yao2024tree} that both models share, which guides them to recognize the objects' states and plan the objects' destinations, grasping types, and placing types.

\section{Experiments}
To quantitatively evaluate the performance of the two proposed methods, we formulate tabletop scenarios where the task is to clear the table (in Sec.~\ref{sec: formulation}). We contribute a multimodal benchmark dataset according to these scenarios that take object states into consideration (in Sec.~\ref{sec: benchmark dataset}). We then evaluate our proposed methods on the dataset (in Sec.~\ref{sec: results}).

\subsection{Task Formulation}\label{sec: formulation}
In our study, we identify two types of object states about leftovers: ``containing leftover food'', where containers like plates or bowls hold liquid or semi-fluid contents that are not directly manipulable by the robot (e.g., a bowl filled with leftover soup), and ``leftover food'', referring to food remains that have been sliced or peeled, which robots can handle.

According to these specified leftover types, we consider three different common scenarios: T1) The instruction is \enquote{clear the table} without specifying what to do with the leftover food. In this scenario, the robot is supposed to generate a manipulation plan for all the objects except those classified as ``leftover food'' or ``containing leftover food''.
In this case, the expected behavior of the robot is to ask the user for handling specifications about the leftovers.
T2) The instruction is \enquote{clear the table and keep all the leftover food}. 
In this scenario, if the object state is ``leftover food'' or ``containing leftover food'', the robot should store it in the fridge. 
T3) The instruction is \enquote{clear the table and discard all the leftover food}. 
In this scenario, if the object state is ``leftover food'', the robot is supposed to throw it into the trash bin. 
However, if the object state is ``containing leftover food'', the robot is supposed to grab the container (not the soup directly) and pour the contents into the trash bin before putting the container into the dishwasher.

\subsection{Benchmark Dataset}\label{sec: benchmark dataset}
\begin{figure}[htbp]
    \begin{subfigure}[b]{0.5\columnwidth}
        \centering
        \fbox{\includegraphics[width=.87\linewidth]{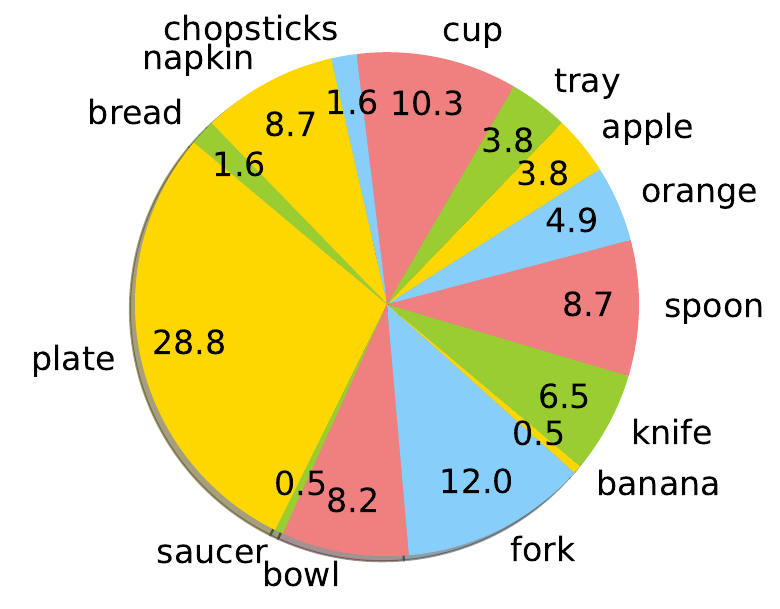}}
         \caption{Distribution of Objects(\%).}
         \label{fig: distribution of objects}
        \centering
    \end{subfigure}
    \hfill
    \begin{subfigure}[b]{0.5\columnwidth}
        \centering
        \fbox{\includegraphics[width=0.95\linewidth]{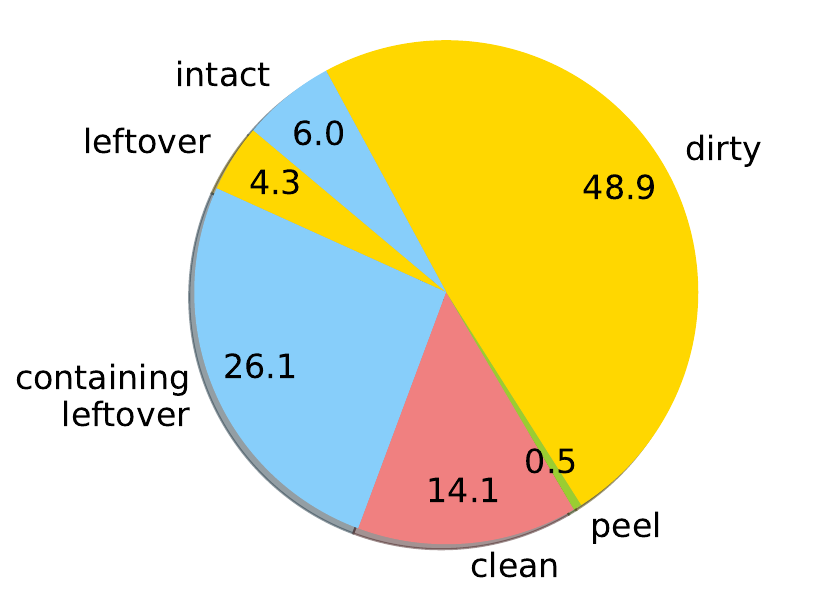}}
        \caption{Distribution of Object States(\%).}
        \label{fig: distribution of object states}
    \end{subfigure}
    \caption{Dataset Statistics}
    \label{fig: dataset statistics}
\end{figure}

To quantitatively evaluate the methods that we propose, we built a benchmark dataset, which is composed of 40 scenes involving 184 objects.
First, we sourced 27 scenes from our daily lives. Second, to create balanced data, we use a diffusion model \cite{rombach2022high} to generate an additional 13 scenes.
Fig.~\ref{fig: distribution of objects} shows all the objects that are involved in this dataset and the proportion among them. The same object may be in different states at different times. The object states that we consider, and their proportion are shown in Fig.~\ref{fig: distribution of object states}.

\subsubsection{Annotation Rules}
The object manipulation plan format, outlined in Sec.~\ref{sec: object manipulation plan}, was employed for annotating the dataset. We asked two individuals to label the dataset according to set rules to reduce bias. 
In the end, they collaborated to resolve discrepancies and finalize the annotations.\\
\textbf{Object name}: use the object name as a JSON key. When more than one item from the same object category is in the scene, add the number behind the name (e.g., `plate 1', `plate 2');
\textbf{Color}: the object color (e.g., `white', `silver', `orange', `red');
\textbf{Size}: the object size, `small', `medium', `big';
\textbf{Shape}: the object shape, `elongated', `irregular', `oval', `round', `spherical', `cylindrical', `rectangle';
\textbf{Container}: use `true' or `false' to label the object as a container or not;
\textbf{State}: If the object is a container, we label it with three states: `clean', `dirty', or 'containing leftover food'. If the object is not a container but edible, we label it with three states: `intact', 'peel', or `leftover food'. If the object is not a container and inedible (fork, knife, spoon), we label it with two states: `dirty' or `clean';
\textbf{Destination}: we use four places: `trash bin', `fridge', `cupboard', and `dishwasher';
\textbf{Grasping type}: we set two types of grasp action: `top grasp' or `edge grasp';
\textbf{Placing type}: In most cases, the robot places the object in the destination. But in the special case that the object is a container that contains leftover food, the robot should pour the leftover food into the trash bin. 

\subsubsection{Evaluation}
In this study, we aim to generate a manipulation plan for the objects to the robot's low-level executor.
The main evaluation metric is accuracy.
There are five parts of the output needed for low-level execution. 
We evaluate our proposed methods' performance in those five parts: 
1) State Detection Accuracy (\textbf{StaA}), 
2) Ambiguous Detection Accuracy (\textbf{AmbA}), 
3) Destination Generation Accuracy (\textbf{DesA}),
4) Grasping Type Generation Accuracy (\textbf{GraA}), and  
5) Placing Type Generation Accuracy (\textbf{PlaA}).
Finally, we calculate one overall accuracy, representing how many objects are being predicted correctly, Completion Accuracy (\textbf{ComA}).

\subsection{Results}\label{sec: results}
First, we test the performance of our proposed methods on object state detection. 
For the modular method, we prompt the GPV-4V(ision)\footnote{gpt-4-vision-preview} as a dense captioning model to get the description of the image. GRiT is another dense captioning model that we used in our experiment.
We prompt an LLM (GPT-4\footnote{gpt-4-0125-preview}) to process the image description (text) to abstract the object states.
For the monolithic method, we prompt GPV-4V(ision) as a vision-language model to directly detect the object state. 
We prompt the pre-trained model in zero-shot or few-shot settings, respectively. 
Therefore, we evaluate six variants: OSSA-LLM-GRiT with the zero-shot setting, OSSA-LLM-GRiT with the few-shot setting, OSSA-LLM-GPT-4V with the zero-shot setting, OSSA-LLM-GRT-4V with the few-shot setting, OSSA-VLM with the zero-shot setting, and OSSA-VLM with the few-shot setting.
\begin{table}[t]
    \caption{Object State Detection Average Accuracy(\%)\scriptsize$\pm$Standard deviation}
    \begin{center}
        \begin{tabular}{c|c}
          \hline
          Method&Result\\
          \hline
          OSSA-LLM(Zero-shot)-GRiT&55.61\scriptsize$\pm$5.05\\
          \hline
          OSSA-LLM((Few-shot)-GRiT&55.69\scriptsize$\pm$4.76\\
          \hline
          OSSA-LLM((Zero-shot)-GPT-4V&73.47\scriptsize$\pm$1.39\\
          \hline
          OSSA-LLM((Few-shot)-GPT-4V&74.77\scriptsize$\pm$1.37\\
          \hline
          OSSA-VLM (Zero-shot)&\textbf{75.14}\scriptsize$\pm$0.37\\
          \hline
          \rowcolor{lightgray}OSSA-VLM (Few-shot)&\textbf{79.83}\scriptsize$\pm$1.00\\
          \hline
        \end{tabular}
        \label{tab: object state detection}
    \end{center}
\vspace{-0.6cm}
\end{table}

As Tab. \ref{tab: object state detection} shows, the monolithic method with few-shot prompts achieves the highest performance compared to the other five variants. 
Overall, the modular model that combines LLM with GRiT performs worse than the other four variants.
However, the advantage of this model is the robot does not need an extra object detection model to determine the object's location. 
The model that combines LLM with GPT-4V is the most expensive in this experiment because it calls two pre-trained models for one plan generation.
The performance is worse than OSSA-VLM with GPT-4V and does not supply the object's location. 
From the results of OSSA-VLM and OSSA-LLM-GPT-4V, we conclude that GPT-4 performs well as a vision-language model that plans multimodal tasks. 
The potential vision information will be lost after the image is converted to text descriptions. For example, spatial reasoning and object attribute understanding are no longer possible.

Second, besides detecting the object state, we prompt the pre-trained LLM and VLM to generate the object manipulation plan consisting of `grasping type', `destination', and `placing type'.
Based on the previous experiment's results, we conclude that GPT-4 functions more efficiently as a VLM in multimodal tasks.
In this experiment, we test two methods: OSSA-LLM-GRiT and OSSA-VLM.
We also prompt methods in zero-shot or few-shot settings respectively.
Hence, we evaluate the four variants in three tasks which are defined in Sec. \ref{sec: formulation}.

\begin{table}[t]
    \caption{Average Accuracy of Object Manipulation Plan Generation. OSSA-L(Z)-G represents zero-shot OSSA-LLM-GRiT, OSSA-L(F)-G represents few-shot OSSA-LLM-GRiT, OSSA-VLM(Z) represents zero-shot OSSA-VLM, OSSA-VLM(F) represents few-shot OSSA-VLM, State Detection Accuracy (StaA), Destination Generation Accuracy (DesA), Grasping Type Generation Accuracy (GraA), Placing Type Generation Accuracy (PlaA), Completion Accuracy (ComA), Ambiguous Detection Accuracy (AmbA), `-' means that the tasks T2 and T3 do not have ambiguity.}
    \begin{center}
        \scalebox{0.92}{\begin{tabular}{c|c|c|c|c|c|c|c}
          \hline 
          &&\multicolumn{6}{c}{ Average accuracy(\%)\scriptsize$\pm$Standard deviation}\\
          \cline{3-8}
          Task&Method&StaA&AmbA&DesA&GraA&PlaA&ComA\\
          \hline \hline 
    \multirow{3}{*}{T1}&OSSA-L(Z)-G&51.09\scriptsize$\pm$0.94&65.78\scriptsize$\pm$12.54&50.37\scriptsize$\pm$2.62&85.83\scriptsize$\pm$1.44&90.74\scriptsize$\pm$2.72&21.74\scriptsize$\pm$1.09\\
          \cline{2-8}
          &OSSA-L(F)-G&50.54\scriptsize$\pm$0.55&95.30\scriptsize$\pm$0.26&57.00\scriptsize$\pm$1.47&\textbf{92.10}\scriptsize$\pm$1.73&96.41\scriptsize$\pm$0.66&26.27\scriptsize$\pm$0.83\\
          \cline{2-8}
          &OSSA-VLM(Z)&71.10\scriptsize$\pm$1.33&92.97\scriptsize$\pm$2.33&83.38\scriptsize$\pm$0.19&89.51\scriptsize$\pm$0.57&\textbf{97.70}\scriptsize$\pm$0.76&52.91\scriptsize$\pm$1.14\\
          \cline{2-8}
          \rowcolor{lightgray}&OSSA-VLM(F)&\textbf{74.36}\scriptsize$\pm$1.78&\textbf{97.37}\scriptsize$\pm$0.07&\textbf{84.81}\scriptsize$\pm$1.29&83.90\scriptsize$\pm$1.74&93.64\scriptsize$\pm$0.84&\textbf{53.09}\scriptsize$\pm$1.12\\
          \hline \hline
    \multirow{3}{*}{T2}&OSSA-L(Z)-G&50.18\scriptsize$\pm$1.13&$-$&59.20\scriptsize$\pm$2.52&88.45\scriptsize$\pm$0.38&90.31\scriptsize$\pm$3.72&22.65\scriptsize$\pm$0.83\\
          \cline{2-8}
          &OSSA-L(F)-G&50.36\scriptsize$\pm$0.63&$-$&46.76\scriptsize$\pm$3.26&\textbf{92.83}\scriptsize$\pm$2.15&\textbf{97.84}\scriptsize$\pm$0.03&21.38\scriptsize$\pm$1.75\\
          \cline{2-8}
          &OSSA-VLM(Z)&68.67\scriptsize$\pm$1.32&$-$&81.17\scriptsize$\pm$0.90&90.20\scriptsize$\pm$1.50&94.99\scriptsize$\pm$2.73&48.63\scriptsize$\pm$1.91\\
          \cline{2-8}
        \rowcolor{lightgray}&OSSA-VLM(F)&\textbf{72.55}\scriptsize$\pm$0.97&$-$&\textbf{85.59}\scriptsize$\pm$1.01&83.72\scriptsize$\pm$1.64&95.48\scriptsize$\pm$1.34&\textbf{54.18}\scriptsize$\pm$1.11\\
          \hline \hline
    \multirow{3}{*}{T3} &OSSA-L(Z)-G&50.72\scriptsize$\pm$0.83&$-$&57.54\scriptsize$\pm$2.44&84.26\scriptsize$\pm$2.70&92.85\scriptsize$\pm$1.26&22.83\scriptsize$\pm$1.44\\
          \cline{2-8}
        &OSSA-L(F)-G&50.18\scriptsize$\pm$0.31&$-$&58.13\scriptsize$\pm$4.45&\textbf{92.06}\scriptsize$\pm$1.67&90.97\scriptsize$\pm$0.65&25.18\scriptsize$\pm$1.25\\
          \cline{2-8}

          &OSSA-VLM(Z)&70.00\scriptsize$\pm$0.39&$-$&74.80\scriptsize$\pm$2.36&91.43\scriptsize$\pm$2.81&\textbf{94.80}\scriptsize$\pm$1.19&47.26\scriptsize$\pm$1.92\\
          \cline{2-8}
        \rowcolor{lightgray}&OSSA-VLM(F)&\textbf{72.19}\scriptsize$\pm$0.87&$-$&\textbf{77.85}\scriptsize$\pm$2.08&84.88\scriptsize$\pm$0.87&91.19\scriptsize$\pm$1.87&\textbf{47.45}\scriptsize$\pm$0.83\\
        \hline
        \end{tabular}
        }
        \label{tab: command generation}
    \end{center}
\vspace{-0.6cm}
\end{table}

The object state detection accuracies shown in Tab.~\ref{tab: object state detection} and Tab.~\ref{tab: command generation} are different. 
When asking for more reasoning items from the pre-trained models (e.g., LLM or VLM), the model's performance decreases.
Similarly, in certain cases humans perform better when focusing on a single task rather than on multiple tasks.
Overall, in those three tasks, the monolithic method OSSA-VLM performs best in ambiguity detection, destination generation, and completion rate.
The few-shot prompts also enhance the performance of models in ambiguity detection and destination generation items.
We notice outliers in the grasping and placing action generation.
For destination generation (\textbf{DesA}), the few-shot monolithic method performs much better than the modular method in the three tasks. 
For grasping type generation (\textbf{GraA}), the modular model OSSA-LLM-GRiT with few-shot prompts performs best.
We conclude that the pre-trained model reasons from the text better than from the image, which needs to consider the object's size and shape. 
For all the variants in the placing action generation (\textbf{PlaA}), the performance is above 90~\%.
From the completion accuracy (\textbf{ComA}), we can see that the few-shot monolithic method OSSA-VLM performs better than the other methods.

\section{Conclusion}
In this paper, we introduced an object state-sensitive agent (OSSA) that can perceive fine-grained scene information (object name and state, etc.) and generate an appropriate object manipulation plan for the robot's low-level executor according to those visual perceptions.
We proposed two approaches: a modular approach consisting of vision (GRiT, or GPT-4V) and Language modules (GPT-4) and a monolithic approach consisting only of a VLM (GPT-4V).
To quantitatively evaluate the two proposed approaches, we formulated an instruction-following task for robots in which the object state needs to be considered in a tabletop scenario.
We demonstrated that VLM GPV-4V supplies more concrete information than the state-of-the-art dense caption model GRiT.
Consequently, the monolithic approach using GPT-4V performed better than the pipeline-based modular approach consisting of the dense caption model GRiT and the language model GPT-4. 
A limitation of the monolithic approach (GPT-4V) is, however, that it is not trained to generate bounding boxes of objects. 
An additional object detection model is needed to get the locations of objects.

In the future, we will develop a model that can distinguish between objects in different states and also localize their location. Furthermore, we will use our models in real scenarios with real robots, taking into account additional objectives such as cost and time for the creation and execution of object state-sensitive plans.

\subsection*{Acknowledgment}
The authors gratefully acknowledge support from the China Scholarship Council (CSC) and the German Research Foundation DFG under project CML (TRR 169).

\bibliographystyle{splncs04} 
\bibliography{references}

\end{document}